\documentclass[10pt,journal,compsoc]{IEEEtran}
\ifCLASSOPTIONcompsoc
  \usepackage[nocompress]{cite}
\else
  \usepackage{cite}
\fi

\ifCLASSINFOpdf
  \usepackage[pdftex]{graphicx}
\else
  \usepackage[dvips]{graphicx}
\fi
\usepackage{amsmath}
\usepackage{acronym}

\usepackage{booktabs}
\usepackage{multirow}
\usepackage{multicol}

\usepackage{url}
\newcommand\MYhyperrefoptions{bookmarks=true,bookmarksnumbered=true,
pdfpagemode={UseOutlines},plainpages=false,pdfpagelabels=true,
colorlinks=true,linkcolor={black},citecolor={black},urlcolor={black},
pdftitle={Bare Demo of IEEEtran.cls for Computer Society Journals},
pdfsubject={Typesetting},
pdfauthor={Michael D. Shell},
pdfkeywords={Computer Society, IEEEtran, journal, LaTeX, paper,
             template}}
\ifCLASSINFOpdf
\usepackage[\MYhyperrefoptions,pdftex]{hyperref}
\else
\usepackage[\MYhyperrefoptions,breaklinks=true,dvips]{hyperref}
\usepackage{breakurl}
\fi

\begin{document}

\title{Gene Teams are on the Field: \\ %
Evaluation of Variants in Gene-Networks Using High %
Dimensional Modelling}

%
%

\author{Suha~Tuna,~Cagri~Gulec,~Emrah~Yucesan,~Ayse~Cirakoglu,%
~Yelda~Tarkan~Arguden$^*$
\IEEEcompsocitemizethanks{\IEEEcompsocthanksitem S. Tuna is with the Department of %
Computational Science and Engineering, Informatics Institute, Istanbul Technical University,  34469, T\"urkiye.\protect
\IEEEcompsocthanksitem C. Gulec is with the Department of %
Medical Genetics, Istanbul Faculty of Medicine, Istanbul University, 34093, T\"urkiye.\protect
\IEEEcompsocthanksitem E. Yucesan is with the Department of 
Neuroscience, Institute of Neurological Sciences,  
Istanbul University Cerrahpasa, 34098, T\"urkiye.\protect
\IEEEcompsocthanksitem A. Cirakoglu and Y. Tarkan Arguden are with the Department of Medical Biology, Faculty of Medicine,  
Istanbul University Cerrahpasa, 34098, T\"urkiye.\protect
\IEEEcompsocthanksitem $^*$The corresponding author. E-mail: yeldata@iuc.edu.tr
}
\thanks{Manuscript received ..., ...; revised ..., ...}}

%
%

\markboth{IEEE/ACM Transactions on Computational Biology and Bioinformatics}%
{Tuna \MakeLowercase{\textit{et al.}}: Evaluation of Variants in Gene-Networks Using High %
Dimensional Modelling}
%



\IEEEtitleabstractindextext{%
\begin{abstract}
In medical genetics, each genetic variant is evaluated as an independent entity regarding its clinical importance. However, in most complex diseases,  variant combinations in specific gene networks, rather than the presence of a particular single variant, predominates. In the case of complex diseases, disease status can be evaluated by considering the success level of a team of specific variants. We propose a high dimensional modelling based method to analyse all the variants in a gene network together. To evaluate our method, we selected two gene networks, mTOR and TGF-$\beta$. For each pathway, we generated $400$ control and $400$ patient group samples. mTOR and TGF-$\beta$ pathways contain $31$ and 93 genes of varying sizes, respectively. We produced Chaos Game Representation images for each gene sequence to obtain $2$-D binary patterns. These patterns were arranged in succession, and a $3$-D tensor structure was achieved for each gene network. Features for each data sample were acquired by exploiting Enhanced Multivariance Products Representation to $3$-D data. Features were split as training and testing vectors. Training vectors were employed to train a Support Vector Machines classification model. We achieved more than $96\%$ and $99\%$ classification accuracies for mTOR and TGF-$\beta$ networks, respectively, using a limited amount of training samples.
\end{abstract}

}

\maketitle

\IEEEdisplaynontitleabstractindextext

%
\IEEEpeerreviewmaketitle

\section{Introduction}

Recently, in parallel with the development of new technologies in genetics, %
it has become possible to study the human genome holistically.
Previously genes were evaluated %
as single entities -we can call those times as \lq\lq analysis era\rq\rq\ of %
genetics- now, the \lq\lq synthesis era\rq\rq\ is born, in which genes are examined %
as parts of a network made up of the whole genome\cite{barabasi2011,choobdar,hawe}. Albert Lazslo Barabasi %
accounted for this situation as \lq\lq disease phenotype is rarely a consequence of an %
abnormality in a single effector gene product, but reflects various pathobio%
logical processes that interact in a complex network.\rq\rq\cite{barabasi2011}. %
In this remarkable concept, genes that encode proteins %
involved in a pathway or known to be associated with a particular disease are %
considered a \lq\lq gene network\rq\rq. Therefore, gene network/s analysis is %
now more reasonable and comprehensible than examining only single genes or %
pathways. The importance of this approach is evident in understanding the %
biogenesis of polygenic-multifactorial diseases that are commonly observed in %
the population and in which the cumulative effect of many mildly acting genes %
is determinative. Unlike single-gene disorders, in polygenic/multifactorial %
diseases, there is not a singular genetic change (mutation) in a %
single underlying %
gene. In addition to environmental factors, a combination of genetic changes %
called polymorphisms or variants plays a role in the emergence of such diseases%
\cite{barabasi2011,choobdar,hawe,maiorino,menche,fang}.

As an analogy, a gene network may be considered as a \lq\lq team\rq\rq. The success of the team relies on the efficiency of the metabolic pathway that contains proteins encoded by genes that make up the gene network. \lq\lq Team success\rq\rq\ is directly related to all players, not just one.  The performance of any team depends on the harmonious working of its individual players. Individual players of a \lq\lq gene team\rq\rq\ are the specific variants of each one of the genes in the network a person carries. Depending on the efficiency of the variant combination, that individual is either healthy or affected in terms of a specific trait. This combinatorial effect of the genes contributes to %
the mechanism of penetrance and expressivity\cite{fahed,rahit}.
If a person has a \lq\lq marvelous\rq\rq\ variant combination -like a \lq\lq %
dream team\rq\rq\ of genes- then that person will be superior in this trait. When there are compensative genes in the gene network for a disease-causing mutation, then the mutant gene’s deleterious effect can be suppressed, and the phenotype appears normal. On the contrary, when many \lq\lq weak\rq\rq\ variants come together in the network, the %
phenotype could be worse than expected from each of these variants. This is already known as one of the mechanisms of the emergence of polygenic multifactorial traits\cite{crouch,lewis}.

Therefore,  when a gene network is determined, it is desirable to be able to identify the combination of variants in that network. If the differences between the gene network variant combinations among individuals could be determined, then it could %
be possible to foresee the susceptibility of that individual to the related diseases%
\cite{fahed,rahit}. The problem with this approach is the insufficiency of the current techniques to examine a gene network as a team.

Currently Genome-wide association studies (GWAS) techniques are used to detect genomic %
variants that may be responsible for the predisposition to complex diseases. %
These studies enable the determination of the most significant variants in %
terms of the related trait/disease coexistence among the variants commonly found %
in people with a particular trait or disease. Using GWAS and bioinformatics %
methods, defining the gene networks underlying certain %
traits/diseases is possible. In this early days of the %
\lq\lq holistic genetics\rq\rq\ era, a lot of research %
focused on this task %
\cite{barabasi2011,choobdar,hawe,maiorino,menche,fang,wangReview}.

One of the many application areas of the results obtained from GWAS studies is the
prediction of an individual's susceptibility to a certain physical or mental illness %
based on their genetic profile. Polygenic Risk Score (PRS) is the standard method used for %
this purpose, and it relies on the SNPs (Single Nucleotide Polymorphisms) that were %
determined as risky for that particular illness/trait by GWAS studies. %
The weighted total scores of all risk SNPs are calculated %
using the %
effect sizes determined in the GWAS study as the weights of the SNPs. Thus, a person-specific Polygenic Risk Score
is determined. Although PRS is a method that can be used as a biomarker to assess individual
susceptibility to diseases, there are currently some limitations that make its clinical application
difficult. One of these is the fact that GWAS studies are still limited to specific ethnic groups, and sometimes
there are groups with different characteristics even within the same population. Another limitation is
that many phenotypic traits are affected by too many genes (polygenicity). Besides, there is no
consensus on which of the various methods used to calculate PRS is the most appropriate. In
particular, the necessity of finding new strategies to %
overcome the polygenicity problem is emphasized%
\cite{uffelmann2021,wangReview,konuma2021,khera2018,silberstein2021,fang,weng2011,xie2021,zhu2018}. 

Methods such as GWAS are highly effective in identifying variants in genes in a particular
disease-associated pathway that are common to most people with the disease. However, these
methods are insufficient in determining patient-specific combinations of other variants in pathway
genes. Regardless of whether they carry risky variants, clinical differences between individuals with
complex diseases are considered to be the result of patient-specific combinations of variants.
Papadimitriu et al. report a machine learning approach to identify digenic or bilocus variant
combinations\cite{papadimitriou2019}. Nevertheless, it is emphasized that \lq\lq the large number of known
variants gives rise to an immense number of combinations, presenting mathematical, statistical, and
computational challenges\rq\rq\cite{mellerup2017}. Therefore,  with the current techniques, it is not possible
to study the combinatorial effects of more than a few variants, let alone all of them. It is obvious that
new approaches are required to overcome the problem. 

Here, we propose a high dimensional modelling based method to analyse all the variants in a gene network together, applying Chaos Game Representation (CGR) \cite{CGR1,CGR2,CGR3,yang2016} as a pre-processing tool to the sequencing data of the genes in the network, and a statistics-based high dimensional feature extraction technique named Enhanced Multivariance Products Representation (EMPR) \cite{Tunga2010,Tunga2013,Tuna2013,Tuna2020}. Then, Support Vector Machines (SVM) which is a flexible and efficient classification algorithm \cite{SVM} was utilized in order to assign the gene network of an individual based on their sequence variants to control or patient groups. 
To test our %
approach, we created  exemplary mTOR and TGF-$\beta$ sub-networks %
consisting of $31$ and $93$ genes, respectively.

\section{Approach}

The biggest problem in processing variant combinations in gene networks is the amount of sequence data. Therefore, to facilitate analysis, we considered applying %
CGR, a technique to convert $1$-D sequence data into $2$-D pattern form \cite{CGR1,CGR2,CGR3}.
The rationale was that the variants in each sequence data would result in slightly different CGR patterns, and computationally sorting out these pattern differences would be easier than comparing sequences.
Afterwards, we had a $2$-D pattern in hand for each gene in the network that needed to be examined together as a team. To do that, we aligned each of the CGR patterns in succession to create a cube as a $3$-D tensor, which would represent an individual’s gene network as a single entity. Then, we adopted EMPR to decompose this $3$-D array and represent it in terms of less dimensional features with the aim of distinguishing control and patient groups according to their variant combinations \cite{Tuna2020}.

To examine the efficacy and the distinguishing capability of our approach, we generated a data set for two gene networks. These are the mTOR and TGF-$\beta$ pathways, each containing $800$ individual $3$-D tensors after applying CGR and aligning the images as a CGR cube.
Half of %
these tensors stand for the control, while the other half denotes the patient groups. %
We split both groups into training and testing parts. Then, we fed the SVM binary %
classification algorithm with three EMPR vector components of the training data and %
generated the learning model. Finally, we calculated the overall accuracy by predicting %
the class (control/patient) of each testing feature according to the constructed SVM model \cite{SVM}.  

\section{Methods}

\subsection{Data Source and Recruitment}

The mTOR\cite{mtor_ref} and TGF-$\beta$\cite{tgf_ref} pathway genes were selected based on the KEGG database %
(\url{https://www.genome.jp/kegg/}) \cite{kegg}. Genomic sequences of the pathway %
genes were fetched from GRCh37 human genome database based on their %
genomic coordinates recorded in the NCBI database %
(\url{https://www.ncbi.nlm.nih.gov/projects/genome/guide/human/index.shtml}). %

\begin{figure}[h!]
\begin{center}
\includegraphics[width=.48\textwidth]{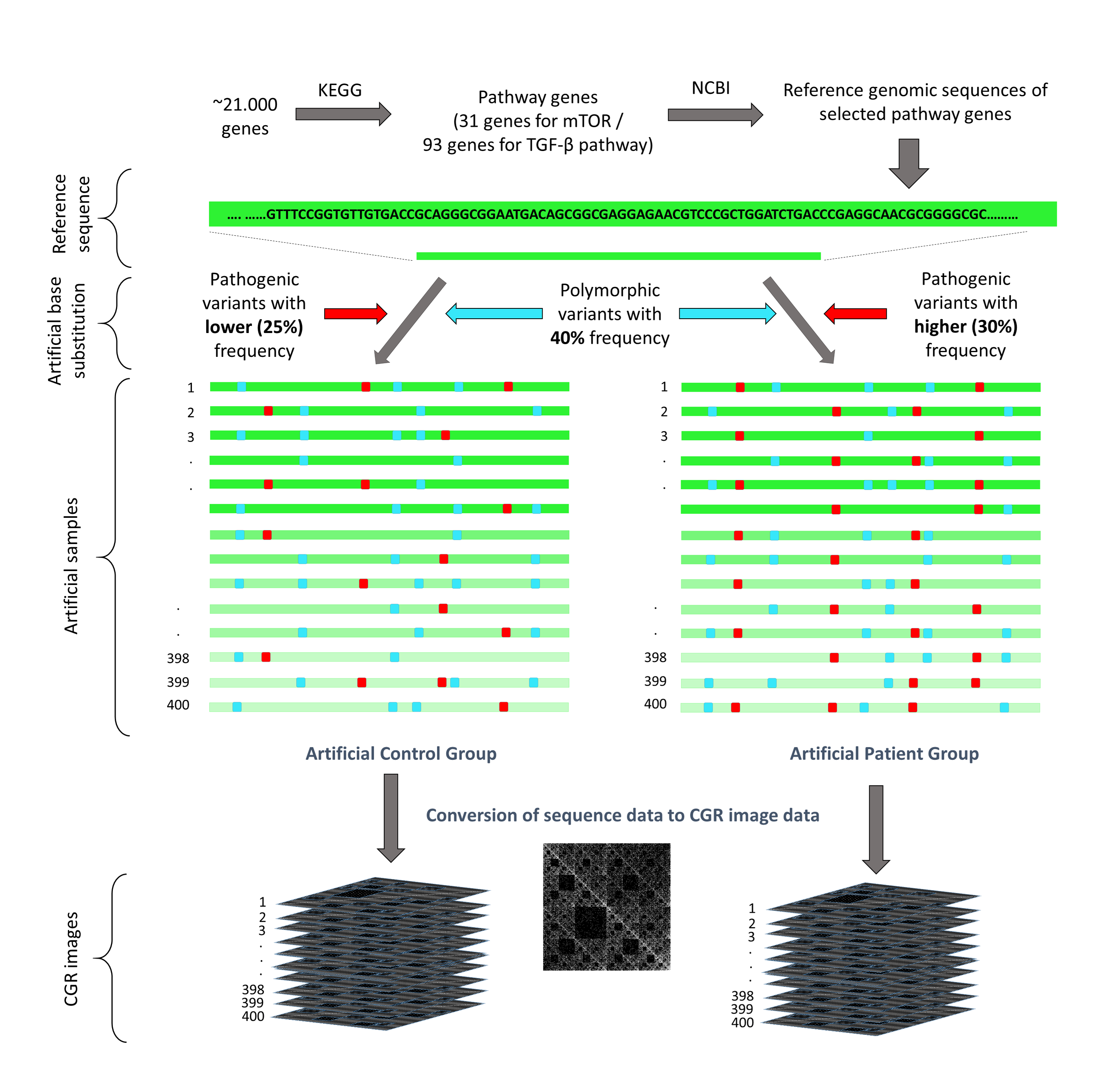}
\end{center}
\caption{Fetching and pre-processing the genomic sequence data}
\label{fig:sequence}
\end{figure}

As represented in Fig. \ref{fig:sequence}, reference sequences composed of each %
gene sequence were %
used as a template to generate $400$ control and $400$ patient sequences for %
each pathway. In the first step, we created two lists of integers for both groups %
that represent the positions of polymorphic and pathogenic variants (\lq polymorphic %
positions list\rq and \lq pathogenic positions list\rq). Each integer in these lists %
has been randomly chosen to be within certain consecutive intervals and exclusive to the %
other list. This interval has been set to $100$ and $200$ for polymorphic and pathogen%
ic variants, respectively (Any integer within the range $1$-$100$, $100$-$200$, $200$-%
$300$, and so on, for \lq polymorphic positions list\rq, and any integer within the %
range $1$-$200$, $200$-$400$, $400$-$600$, and so on, for \lq pathogenic positions list%
\rq). In the second step, the reference base at each position represented in the \lq polymorphic %
positions list\rq\ was replaced by the variant base in $40\%$ of both control and patient %
sequences. The alterations in these positions were accepted as non-pathogenic and/or %
common variants with $0.40$ minor allele frequency in both groups. In the next step, the %
reference base at each position represented in the \lq pathogenic positions list\rq was %
replaced by the variant base in $25\%$ of control sequences and $30\%$ of patient %
sequences. The alterations in these positions were accepted as disease-associated/pathogen%
ic variants with $0.25$ allele frequency in the control group and $0.30$ allele frequen%
cy in the patient group. In all these steps, we set minor allele frequency (MAF) higher %
because, contrary to single-gene disorders where rare variants (with MAF$<0.01$) %
are causative, complex disorders are the consequences of the combination of the varian%
ts with higher allele frequency (MAF$>0.01$). All variant sequences were %
in the haploid state. The properties of the datasets are %
summarised in Supp. Table 1 and Supp. Table 2.
\vskip-.4cm

The known available datasets, e.g., 1000 Genomes, GENESIS, Solve-RD, %
Munich Exome (EVAdB), Baylor-Hopkins Center for Mendelian Genomics (BH-CMG), 100KGP, %
GeneDx, and NHLBI-GO Exome Sequencing Project (ESP) databases, have not used preferably %
to avoid any bias (it is difficult to distinguish patient from control dataset). %
Therefore, we created datasets that we arranged according to the percentage of the allele frequency. Since real human samples or data were not used in the study, ethics committee approval was not considered necessary.

To evaluate the efficiency of the proposed method, both control and patient %
groups belonging to each pathway dataset were split into two independent and non-intersecting parts. %
The first part was considered the training, %
while the latter was called the testing data. These separate subsets for each %
pathway dataset were symbolised as %
$D_{\text{train}}$ and $D_{\text{test}}$, respectively. $D_{\text{train}}$ was collected %
by generating randomly selected pathways among $400$ control and $400$ patient networks at %
a certain amount. For the classification phase, the number %
of elements in $D_{\text{train}}$ %
was assumed to be less than the number of networks in $D_{\text{test}}$. %
$D_{\text{train}}$ was utilised for training %
the classification algorithm, while %
$D_{\text{test}}$ was employed to verify the efficacy of the training model. To provide a %
convenient learning model and determine whether a given network in $D_{\text{test}}$ %
belongs to the control or the patient class, we applied a new feature extraction %
approach based on CGR and EMPR. %

\subsection{Chaos Game Representation}

\begin{figure}[h!]
\centerline{\includegraphics[width=.48\textwidth]{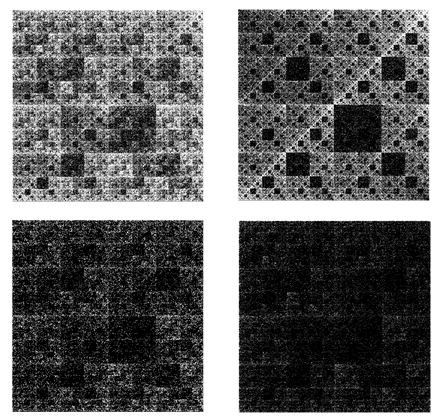}}
\caption{$700\times 700$ CGR images corresponding to four genes in mTOR and TGF-$\beta$ pathways: %
RPTOR of mTOR (top-left), GSK3B of mTOR (top-right), SMAD6 of %
TGF-$\beta$ (bottom-left), SMAD7 of TGF-$\beta$ (bottom-right)}
\label{fig:CGRimages}
\end{figure}

CGR is an efficient technique that converts long $1$-D genomic sequences into $2$-D %
images (see Fig. \ref{fig:CGRimages}), say patterns \cite{CGR1,CGR3,CGR2}. In this %
manner, CGR enables to pull of %
significant data parts out from the corresponding gene sequence using a convenient %
feature extraction method suitable for images. %

In the DNA sequence case, the corresponding CGR of a sequence is nothing but a square-shaped %
binary image whose bottom-left corner overlaps with the origin of  %
$2$-D Cartesian space. If Adenine is assumed to be depicted with the origin, which is %
the point $(0,0)$, Cytosine is placed at the point $(0,1)$, Guanine is located at %
$(1,0)$, while Thymine stands at the final corner, that is $(1,1)$. The pattern is %
initialized with a point on the centre in the image, that is $(0.5,0.5)$. The first %
point of the pattern is settled in the half way between the centre and the corner cor%
responding to the first nucleotide of the sequence. In general, the $i$-th point of the %
image is then placed just in the middle of the $(i-1)$-th point and the vertex cor%
responding to the $i$-th nucleotide. %

Formally, if the horizontal and the vertical coordinates of the $i$-th nucleotide %
of a given sequence are defined as $X_i$ and $Y_i$, respectively, these entities %
are determined using the following linear equations %
\begin{align}
X_i&=\frac{1}{2}\left(X_{i-1}+C_i^{(x)}\right) \nonumber \\
Y_i&=\frac{1}{2}\left(Y_{i-1}+C_i^{(y)}\right)
\label{eq:CGRiters}
\end{align}
where $X_0=Y_0=0.5$. In (\ref{eq:CGRiters}), $C_i^{(x)}$ and $C_i^{(y)}$ stand %
for the coordinates of the pre-defined corners of the unit-square, that is %
$[\,0,1\,]^2$, related to the corresponding nucleotide mentioned above. %

The resolution of the CGR image is adjustable and may affect the representation %
quality of the gene sequence under consideration. For instance, if the size of %
the CGR image is selected too small, then some of the points can overlap, and this %
fact can prevent the contribution of the overlapping points to the whole pattern. %
On the other hand, in case the size of the image is selected too large, some %
unnecessary gaps between the points may occur and the representation eligibility %
of the CGR pattern is influenced negatively. Thus, fixing the optimal resolution %
for a CGR image is also crucial to improving %
the representation quality. %

To process the pathways under consideration as a whole %
and extract meaningful features using Enhanced Multivariance Products Representation, %
all CGR images of the genes in the pathways are aligned in succession. %
Thus, a $3$-D representation for any individual mTOR or TGF-$\beta$ %
gene network %
is constructed. The emerged $3$-D data is named as \emph{CGR cube} of a gene network %
and is suitable for processing by the proposed high-%
dimensional modelling method. %

\subsection{Enhanced Multivariance Products Representation}
\label{subsec:EMPR}

Enhanced Multivariance Products Representation (EMPR) is a high dimen%
sional data decomposition method \cite{Tunga2010,Tunga2013,Tuna2013,Tuna2020}. %
It enables a representation of multidimensional data in terms of %
lower-dimensional %
entities. Accordingly, EMPR can be considered as a finite series of lower %
dimensional components. This aspect of EMPR enables to reduce the dimensionality %
of multidimensional data and simplifies further analysis.

In scientific experiments and applications, one of the crucial challenges %
in analysing data is the \lq\lq curse of dimensionality\rq\rq\ \cite{curseDim}. %
Therefore, governing this issue by reducing the number of dimensions %
becomes critical. Thus, EMPR can be regarded as a suitable %
technique for addressing multidimensional problems. %

EMPR is an extension of a well-known statistical method %
called High Dimensional Model Representation (HDMR) \cite{sobol,alis1999}. HDMR was %
invented for decomposing and decorrelating the inputs in multidim%
ensional input-output systems \cite{sobol}. In a general multidimensional system, %
each input, say dimension, contributes to the behaviour of the output individually %
or cooperatively with other inputs \cite{alis1999,alis2001,rabitz}. However, %
determining these contributions is significant to evaluate the corresponding %
model for meta-modelling \cite{ayresMeta,kubicekSurrogate}, sensitivity analysis %
\cite{liuSensitivity} and reduction \cite{chowdhuryReduction}, %
etc. %

As HDMR, EMPR is capable of dealing with $N$-D data. But in this study, %
the $3$-D case is considered without loss of generality. However, all %
formulations which will be presented here can be generalised to the $N$-D %
case without any difficulty. Further in this section, EMPR for Gene %
Network Analysis (GNA) will be introduced and discussed. %

Let $\mathbf{G}$ denote the $3$-D CGR cube and assume its size is %
$n_1\times n_2\times n_3$. This means the network $\mathbf{G}$ has $n_3$ %
gene sequences, each of which has various sizes and is represented %
through $n_1\times n_2$ binary images, thanks to the CGR method. %
Then, the EMPR expansion of the CGR cube can be explicitly given as follows %
\begin{align}
\label{eq:EMPR}
\mathbf{G}=\;g^{(0)}\left[\bigotimes_{r=1}^3\mathbf{s}^{(r)}\right]+%
\sum_{i=1}^3\mathbf{g}^{(i)}\otimes\left[\bigotimes_{r=1\atop r\neq i}%
^3\mathbf{s}^{(r)}\right]\nonumber \\
+
\sum_{i,j=1\atop i<j}^3\mathbf{g}^{(i,j)}\otimes\left[\bigotimes_%
{r=1\atop r\neq i,j}^3\mathbf{s}^{(r)}\right]+\mathbf{g}^{(1,2,3)}.
\end{align}
In formula (\ref{eq:EMPR}), $g^{(0)}$, $\mathbf{g}^{(i)}$, and $\mathbf{g}^{(i,j)}$ %
denote the zero-way, the one-way, and the two-way \emph{EMPR %
components}, respectively, and $\otimes$ stands for the outer product %
\begin{figure}[h!]
\begin{center}
\includegraphics[width=.48\textwidth]{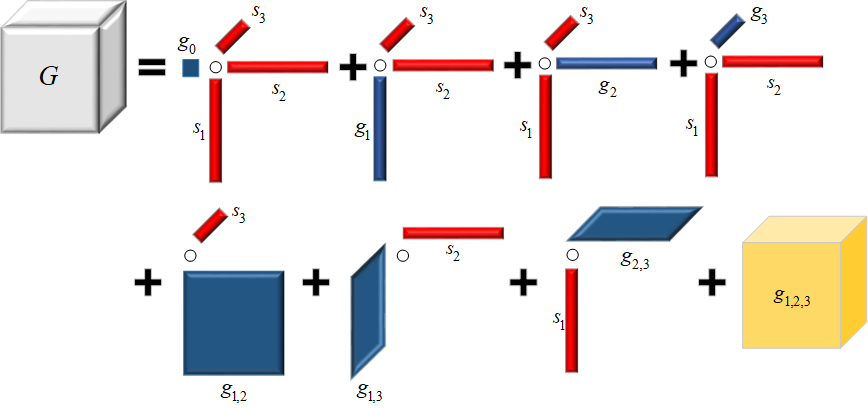}
\end{center}
\caption{Graphical demonstration of EMPR expansion for $3$-D case.}
\label{fig:empr}
\end{figure}
operation \cite{tensor}. The $3$-D EMPR expansion is a finite sum. Thus, %
it involves exactly $2^3$ EMPR components \cite{Tunga2010,Tunga2013,%
Tuna2013,Tuna2020}. The graphical expression %
of the EMPR decomposition is given in Fig. \ref{fig:empr}. %

In (\ref{eq:EMPR}), $g^{(0)}$ is a scalar that can be considered as a %
$0$-D entity. $\mathbf{g}^{(i)}$ stands for $1$-D structures,  which are %
the vectors, and $\mathbf{g}^{(i,j)}$ %
denotes the $2$-D entities which can be acknowledged as %
the matrices. Additionally, other entities involved in %
(\ref{eq:EMPR}) and denoted by $\mathbf{s}^{(r)}$ are $1$-D elements 
and called the \emph{support vectors} \cite{Tuna2020}. %
In this sense, $\mathbf{s}^{(r)}$ is the $r$-th support vector %
that resides on the $r$-th axis of the $3$-D CGR cube where %
$r=1,2,3$. Thus, one can easily verify that the $r$-th support vector %
is an entity composed of $n_r$ elements. Support vectors are multiplied %
with the corresponding EMPR components in outer product manner and %
enhance its dimensionality. Besides, they provide flexibility %
for EMPR expansion and must be selected rationally. This choice is crucial %
since it affects the representation eligibility of the EMPR expansion. %

Since EMPR has an additive nature, $\mathbf{G}$ should be expressed in %
terms of $3$-D structures. As a consequence of outer products between EMPR %
components and support vectors, new $3$-D but less complicated entities %
are established. These new elements are called \emph{EMPR terms} %
\cite{Tunga2010, Tunga2013, Tuna2013, Tuna2020}. Each EMPR term is %
named regarding its EMPR component. Thus, the term constructed with %
$g^{(0)}$ and all three supports are called the \emph{zeroth EMPR term}. %
The term composed of $\mathbf{g}^{(i)}$ and the remaining two support %
vectors (except the $i$-th one) is called the \emph{$i$-th EMPR term}. Similarly, %
the term including $\mathbf{g}^{(i,j)}$ and the corresponding support %
vector are called \emph{$(i,j)$-th EMPR term}. It is clear that %
all EMPR terms are of size $n_1\times n_2\times n_3$, just as the original %
data, $\mathbf{G}$. %

Additionally, during the EMPR process, three weight vectors can be exploited %
to adjust the contributions of each CGR pixel in $\mathbf{G}$. The weight %
vectors are consisted of non-negative real values and must satisfy the %
following conditions %
\begin{equation}
\label{eq:weights}
\left\|\boldsymbol{\omega}^{(1)}\right\|_1=1,\quad
\left\|\boldsymbol{\omega}^{(2)}\right\|_1=1,\quad
\left\|\boldsymbol{\omega}^{(3)}\right\|_1=1.
\end{equation}
In (\ref{eq:weights}), it is clear that the sum of all elements %
for each weight vector should be equal to $1$. These equations hold %
due to the statistical necessities, and they facilitate the comp%
utations in the evaluation process of EMPR components. %

However, the EMPR components should satisfy the following constraints %
\begin{equation}
\label{eq:vanish}
\sum_{i_p=1}^{n_p}\boldsymbol{\omega}^{(p)}_{i_p}\,\mathbf{s}^{(p)}_{i_p}\,%
\mathbf{g}_{i_1,\ldots,i_m}^{(1,\ldots,m)}=0;\quad 1\leq p\leq m\in\{1,2,3\}%
\end{equation}

where $\mathbf{s}^{(p)}_{i_p}$ and $\boldsymbol{\omega}^{(p)}_{i_p}$ %
are the $i_p$-th elements of the $p$-th support vector and %
$p$-th weight vector, respectively. However, 
$\mathbf{g}_{i_1,\ldots,i_m}^{(1,\ldots,m)}$ stands for the %
$(i_1,\ldots,i_m)$-th entry of the corresponding EMPR component %
$\mathbf{g}^{(1,\ldots,m)}$. The equalities in (\ref{eq:vanish}) %
are called \emph{vanishing conditions}. They lead to two essential %
properties of EMPR components, which are the uniqueness under a %
certain set of support vectors and the mutual orthogonality. %

By employing the vanishing conditions in (\ref{eq:vanish}) and adopting %
the weight vectors given in (\ref{eq:weights}) with the pre-selected %
support vectors, the scalar EMPR component, i.e. $g^{(0)}$, can be %
determined uniquely as follows %
\begin{equation}
\label{eq:0thcomp}
g^{(0)}=\sum_{i=1}^{n_1}\sum_{j=1}^{n_2}\sum_{k=1}^{n_3}%
\boldsymbol{\omega}^{(1)}_i\,\boldsymbol{\omega}^{(2)}_j\,%
\boldsymbol{\omega}^{(3)}_k\,\mathbf{s}_i^{(1)}\,\mathbf{s}_j^{(2)}%
\,\mathbf{s}_k^{(3)}\,\mathbf{G}_{ijk}.%
\end{equation}
It is possible to mark that the right-hand side of the equation %
(\ref{eq:0thcomp}) denotes a weighted sum of $\mathbf{G}$ multiplied %
by the relevant support vector elements over all axes. Thus, the %
zero-way EMPR component associates with a specific weighted average value of %
the CGR cube, $\mathbf{G}$. %

If the conditions in (\ref{eq:weights}) and constraints (\ref{eq:vanish}) %
are exploited again, the elements of three one-way EMPR components are %
calculated uniquely as follows %
\begin{align}
\label{eq:jthcomp}
\mathbf{g}_i^{(1)}&=\sum_{j=1}^{n_2}\sum_{k=1}^{n_3}%
\boldsymbol{\omega}^{(2)}_j\,%
\boldsymbol{\omega}^{(3)}_k\,\mathbf{s}_j^{(2)}\,\mathbf{s}_k^{(3)}\,%
\mathbf{G}_{ijk}-g^{(0)}\,\mathbf{s}_i^{(1)}, \nonumber \\
\mathbf{g}_j^{(2)}&=\sum_{i=1}^{n_1}\sum_{k=1}^{n_3}%
\boldsymbol{\omega}^{(1)}_i\,%
\boldsymbol{\omega}^{(3)}_k\,\mathbf{s}_i^{(1)}\,\mathbf{s}_k^{(3)}\,%
\mathbf{G}_{ijk}-g^{(0)}\,\mathbf{s}_j^{(2)}, \nonumber \\
\mathbf{g}_k^{(3)}&=\sum_{i=1}^{n_1}\sum_{j=1}^{n_2}%
\boldsymbol{\omega}^{(1)}_i\,%
\boldsymbol{\omega}^{(2)}_j\,\mathbf{s}_i^{(1)}\,\mathbf{s}_j^{(2)}\,%
\mathbf{G}_{ijk}-g^{(0)}\,\mathbf{s}_k^{(3)}. %
\end{align}
while the rest of the components can be computed in a similar manner. %

As addressed, the components $\mathbf{g}^{(1)}$, %
$\mathbf{g}^{(2)}$, and $\mathbf{g}^{(3)}$ are one-way entities. %
Therefore, each forms a vector lying on its corresponding axis. %
According to (\ref{eq:0thcomp}) and (\ref{eq:jthcomp}), $\mathbf{g}^{(1)}$ %
is obtained by squeezing the CGR cube through its front and upper sides, %
respectively. $\mathbf{g}^{(2)}$ is obtained by suppressing the CGR cube %
through its front and right sides. The last vector, that is %
$\mathbf{g}^{(3)}$, is evaluated by compressing the cube through its upper %
and right sides. After these suppression steps, the means associated with %
certain dimensions are procured. Then, the relevant support vector %
weighted with $g^{(0)}$ is subtracted from the calculated mean. %
Thus, each one-way EMPR term defines the attitude and %
individual contribution of the corresponding dimension (axis) to the whole %
network $\mathbf{G}$. In this sense, $\mathbf{g}^{(1)}$ and %
$\mathbf{g}^{(2)}$ terms specify both dimensions of the surrogate %
CGR pattern emerged from $\mathbf{G}$. This CGR pattern is a weighted %
average of CGR images belonging to all genes in the corresponding net%
work. However, the third one-way EMPR term, %
$\mathbf{g}^{(3)}$, interprets the interrelation among the CGR images %
of the genes of the network. Thus, each one-way EMPR term characterizes %
the $\mathbf{G}$ in its own way and can be exploited as low dimensional %
features for the $3$-D gene network data on the focus.

Finally, in this section, we will provide the details about the properties %
and selection process of the EMPR support vectors. As a beginning, the %
support vectors should satisfy the following normalization conditions 
\begin{equation}
\label{eq:supnorm}
\sum_{i_p=1}^{n_p}\boldsymbol{\omega}^{(p)}_{i_p}\,%
\left[\mathbf{s}^{(p)}_{i_p}\right]^2=1;\qquad p=1,2,3.%
\end{equation}
under the given weight vectors. With the help of the conditions in %
(\ref{eq:supnorm}), the support vectors can be selected independently from %
the magnitude. Thus, each support vector indicates the relevant %
direction where it acts as a weight vector to the contributions which are %
stored as the elements of EMPR components. %

Any suitable set of vectors can be employed as the support vector team %
for EMPR, as long as they are in harmony with the conditions in %
(\ref{eq:vanish}) and (\ref{eq:supnorm}). For this reason, the vectors %
whose elements are given explicitly as %
\begin{align}
\label{eq:ADS}
\mathcal{S}_i^{(1)}&=\sum_{j=1}^{n_2}\sum_{k=1}^{n_3}%
\boldsymbol{\omega}_j^{(2)}\,\boldsymbol{\omega}_k^{(3)}\,%
\mathbf{G}_{ijk}, \nonumber \\
\mathcal{S}_j^{(2)}&=\sum_{i=1}^{n_1}\sum_{k=1}^{n_3}%
\boldsymbol{\omega}_i^{(1)}\,\boldsymbol{\omega}_k^{(3)}\,%
\mathbf{G}_{ijk}, \nonumber \\
\mathcal{S}_k^{(3)}&=\sum_{i=1}^{n_1}\sum_{j=1}^{n_2}%
\boldsymbol{\omega}_i^{(1)}\,\boldsymbol{\omega}_j^{(2)}\,%
\mathbf{G}_{ijk}. %
\end{align}
can be adapted as the support vectors of an EMPR expansion, after %
performing normalisation according to (\ref{eq:supnorm}). %

The support vectors in (\ref{eq:ADS}) can be calculated in a %
straightforward manner and exploited in EMPR expansion as long as %
they do not vanish \cite{Tunga2010,Tuna2020}. From (\ref{eq:ADS}), %
it is obvious that each formula denotes a weighted average of the %
CGR cube $\mathbf{G}$ over all axes but the one direction (axis). %
Thereby, the equations in (\ref{eq:ADS}) %
indicate averaged directions for the CGR cube. To %
this end, these support vectors in (\ref{eq:ADS}) are called %
\emph{Averaged Directional Supports (ADS)} \cite{Tuna2020} and can %
be encountered in several EMPR applications existing in the scientific %
literature \cite{Tunga2010,Tunga2013,Tuna2013,Tuna2020}. %
In this study, the ADS are employed in order to %
extract features using EMPR. However, the constant weight vectors whose %
elements are as follows %
\begin{equation}
\boldsymbol{\omega}_i^{(1)}=\frac{1}{n_1},\quad 
\boldsymbol{\omega}_j^{(2)}=\frac{1}{n_2},\quad
\boldsymbol{\omega}_k^{(3)}=\frac{1}{n_3}
\end{equation}
will be exploited as the weights in EMPR processes.

In summary, EMPR enables to extract features from $3$-D CGR cubes. %
These features are the vector EMPR components given in %
(\ref{eq:jthcomp}). The vectors are ensembled to form a long %
feature vector. Each of these vectors spans all dimensions of the %
CGR cube under consideration with one accord. Therefore, %
the Support Vector Machines algo%
rithm can be fed with the ensembled feature vectors, and an %
efficient learning model can be constructed. %

\subsection{Support Vector Machines}

Determining whether a given gene network belongs to the patient or the control group %
is the main aim of the present work. Thus, extracting practical and meaningful features and selecting an appropriate %
classifier that is in harmony with these features are crucial. Since data classification is one of the major challenges in %
machine learning, many techniques are proposed both for supervised and unsupervised cases. %
Support Vector Machines (SVM), a flexible supervised classification algorithm, is considered %
as an effective technique for grouping pre-labeled data \cite{SVM}. 
The aim of SVM is to construct a hyperplane whose margins with each cumulated point %
set (class) are the widest possible. If the collected data points are overlayed %
separate enough, then it becomes possible to distinguish them into homogeneous %
groups using a linear hyperplane (or linear kernel). Otherwise, a non-linear %
kernel should be exploited to obtain a satisfactory classification accuracy. This approach is called \emph{the kernel trick} \cite{SVMkernel}.%

The main aim of this study is to determine whether a given gene network %
belongs to the control or patient group. Thus, we formulate this problem as a %
binary classification task. %
To classify the data in $D_{\text{test}}$, first, the SVM model should be trained using %
$D_{\text{train}}$. The elements of $D_{\text{train}}$ and $D_{\text{test}}$ are %
CGR cubes defined in subsection 3.2 are $3$-D. Thus, it is hard to train the model by %
feeding SVM with the CGR cubes. To overcome this fact, the SVM %
algorithm is trained with the vector EMPR components of %
each CGR cube whose explicit formulae are given in (\ref{eq:jthcomp}). %
Therefore, a feature vector for each CGR cube is constructed by ensembling %
the one-way EMPR components corresponding to the CGR cube as follows %
\begin{equation}
\mathbf{f}=\left[\,{\mathbf{g}^{(1)}}^T\quad {\mathbf{g}^{(2)}}^T%
\quad {\mathbf{g}^{(3)}}^T\,\right]^T.
\end{equation}
If the CGR cubes are generated as the size of $n_1\times n_2\times n_3$, then the %
length of each %
feature vector $\mathbf{f}$ becomes $n_1+n_2+n_3$. This means the hypersurface %
created by the SVM algorithm lays in $n_1+n_2+n_3$ dimensional space. Though this %
number may seem quite large, the features whose %
distinguishing capabilities are satisfactory may reduce the computation %
complexity of SVM significantly. %

To train the SVM model, $\mathbf f$ features of the CGR cubes in %
$D_{\text{train}}$ are evaluated. Then, the SVM model is trained using %
these feature vectors. After the training phase, $\mathbf f$ features %
of the CGR cubes in $D_{\text{test}}$ are %
given to the trained model, and %
the class of each feature which belongs to $D_{\text{test}}$ is %
predicted. Consequently, the statistics for the objective evaluation %
of the proposed estimator are calculated using the elements of the %
corresponding confusion matrix obtained in each independent run.

\section{Results}

In this section, we will provide the results obtained by assembling %
CGR, EMPR, and SVM for the mTOR and TGF-$\beta$ gene network datasets. %
To this end, we performed several computational efforts to emphasise %
the efficiency of the proposed method. Since the aim of this study is %
to present an efficient classification method for the gene pathways, %
the overall accuracy (OA) is considered as the fundamental %
objective assessment metric. The OA value for each experiment is calculated %
as follows %
\begin{equation}
\text{OA}=\frac{\text{Number of correct predictions}}%
{\text{Number of testing samples}}\times 100.
\end{equation}
However, since OA could yield limited information about the %
classifier performance, we also reported the true negative rate, true positive %
rate (precision), recall (sensitivity), specificity, and Matthew's Correlation %
Coefficient (MCC) metrics \cite{MLmetrics,mcc}. %
The reported statistics are the average of $100$ independent SVM %
runs. %
Before the training stage, all features belonging to the training and %
the testing set were normalised. In the SVM phase, we adopted Radial Basis Function %
(RBF) kernel as the SVM %
kernel. To determine the best classifier parameters $c$ and $\gamma$, which %
controls the behaviour of the RBF kernel, we performed %
a $5$-fold cross-valida%
tion and grid search on a %
$9\times 9$ grid $[10^{-4},\, 10^{-3},\, \ldots,\, 1,\, %
\ldots,\,10^{3},\, 10^{4}]\times[10^{-4},\, 10^{-3},\, \ldots,\, 1,\, %
\ldots,\,10^{3},\, 10^{4}]$. Finally, the model was trained using an SVM %
algorithm implemented by the LIBSVM package \cite{libSVM}. 
\begin{figure}[h!]
\begin{center}
\includegraphics[width=.5\textwidth]{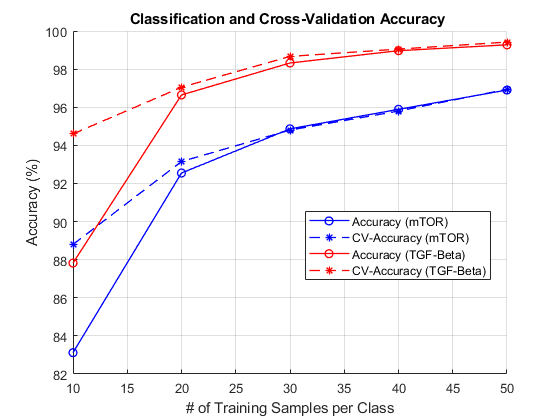}
\end{center}
\caption{Average overall and cross-validation accuracies for varying %
training sample counts.}
\label{fig:accuracy}
\end{figure}
In Fig. \ref{fig:accuracy}, we provided the classification and cross-validation %
accuracies for both gene pathway datasets. We performed the trials for %
various training sample amounts both for control and patient groups. These %
amounts vary between $10$ and $50$ with an increment of $10$. After resolving %
the number of training samples for each class, the fixed number of %
training samples were selected randomly. Then, the rest of the networks in %
each dataset were reserved for testing. %

It is clear from Fig. \ref{fig:accuracy} that the proposed method %
yields higher than $90\%$ classification accuracy using only $20$ training %
samples for both mTOR and TGF-$\beta$ datasets. Initially, the OA values for %
mTOR and TGF-$\beta$ networks are calculated as about $88\%$ and $83\%$ %
for $10$ training samples from both classes, respectively. Then, these values %
increase to about $97\%$ and $93\%$ rapidly. The increments for both datasets %
are consistent as the number of training samples from control and patient %
classes grows. Furthermore, the cross-validation (CV) accuracies for both %
datasets tend to escalate while the number of training samples increases %
and are in harmony with the observed OA results. It is evident that %
the gap between the corresponding OA and CV accuracy tends to decrease %
consistently both for mTOR and TGF-$\beta$ while the training sample count %
grows, especially after dealing with $20$ training samples. %

\begin{table}[h!]
\caption{Classifier performance metrics for mTOR and TGF-$\beta$ datasets.}
\centering
\begin{tabular*}{.49\textwidth}{@{}llccccc@{}}
\toprule
& \textbf{Metric} / $\mathbf{S}$ & $\mathbf{10}$ & $\mathbf{20}$ & $\mathbf{30}$ & %
$\mathbf{40}$ & $\mathbf{50}$ \\
\midrule
\multirow{5}{*}{\rotatebox[origin=c]{90}{mTOR}}%
& True Neg. Rate & $0.9087$ & $0.9265$ & $0.9398$ & $0.9463$ & $0.9579$ \\
& True Pos. Rate & $0.8127$ & $0.9315$ & $0.9596$ & $0.9738$ & $0.9813$ \\
& Recall         & $0.9011$ & $0.9227$ & $0.9375$ & $0.9438$ & $0.9565$ \\
& Specificity    & $0.7613$ & $0.9282$ & $0.9598$ & $0.9740$ & $0.9816$ \\
& MCC            & $0.6894$ & $0.8544$ & $0.8984$ & $0.9189$ & $0.9386$ \\
\midrule
\multirow{5}{*}{\rotatebox[origin=c]{90}{TGF-$\beta$}}%
& True Neg. Rate & $0.9608$ & $0.9854$ & $0.9902$ & $0.9949$ & $0.9949$ \\
& True Pos. Rate & $0.8407$ & $0.9506$ & $0.9770$ & $0.9846$ & $0.9907$ \\
& Recall         & $0.9589$ & $0.9853$ & $0.9902$ & $0.9949$ & $0.9949$ \\
& Specificity    & $0.7945$ & $0.9476$ & $0.9763$ & $0.9844$ & $0.9906$ \\
& MCC            & $0.7765$ & $0.9344$ & $0.9668$ & $0.9794$ & $0.9855$ \\
\bottomrule
\end{tabular*}
\label{tab:tab1}
\end{table}
After discussing the classification accuracy of the %
suggested method, we also need to evaluate the performance and stability of the %
proposed estimator based on CGR, EMPR, and SVM. To this end, the widely used %
machine learning metrics for the estimator assessment, such as true negative rate, %
true positive rate (precision), recall (sensitivity), specificity, %
and MCC, were provided in Table \ref{tab:tab1}. In Table \ref{tab:tab1}, %
the specified metrics are tabulated for %
increasing training sample counts from control and patient classes for both mTOR %
and TGF-$\beta$ datasets.

It is obvious from Table \ref{tab:tab1} that each metric approaches value %
$1$ consistently as the number of training samples grows. However, the %
True Positive Rate, specificity, and MCC values may be considered a bit low %
at $10$ training samples from control and patient classes for both datasets. %
Nevertheless, these values increased rapidly both for mTOR and TGF-$\beta$ as %
$20$ or more training samples were employed. We can easily verify from %
Table \ref{tab:tab1} %
that all stability metrics are calculated above $0.93$ and $0.98$ for mTOR %
and TGF-$\beta$ datasets, respectively, by exploiting $50$ training samples %
from both control and patient groups. The reported values address that the %
proposed estimator achieves significant success in accurately classifying the %
networks belonging to control and patient samples for the considered mTOR and %
TGF-$\beta$ datasets.
\begin{figure}[h!]
\begin{center}
\includegraphics[width=.5\textwidth]{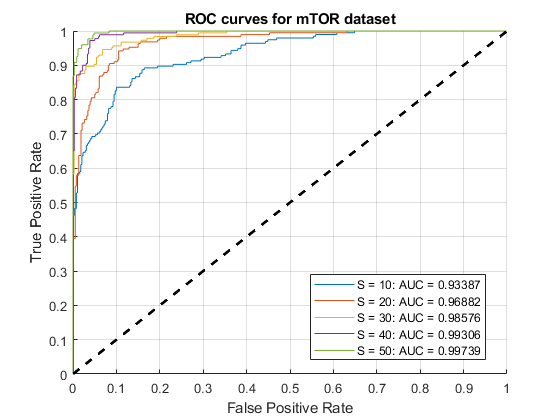}
\end{center}
\caption{ROC curves and AUC values for mTOR dataset with varying training %
sample counts.}
\label{fig:roc_mTOR}
\end{figure}
\begin{figure}[h!]
\begin{center}
\includegraphics[width=.5\textwidth]{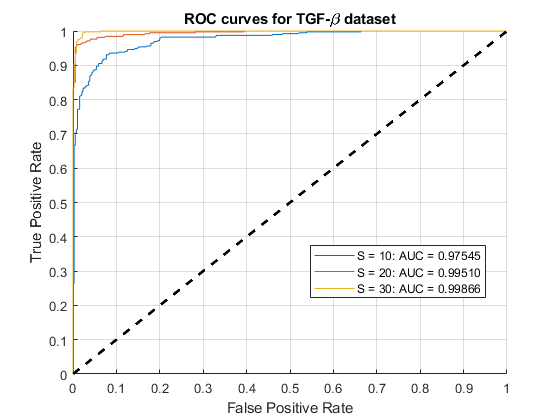}
\end{center}
\caption{ROC curves and AUC values for TGF-$\beta$ dataset with varying %
training sample counts.}
\label{fig:roc_tgf}
\end{figure}

As the further assessment of the proposed CGR, EMPR, and SVM assemble, 
receiver operating characteristic (ROC) curves for both datasets are presented %
in Fig. \ref{fig:roc_mTOR} and \ref{fig:roc_tgf}, where the corresponding %
area under curve (AUC) values are provided %
therein. In Fig. \ref{fig:roc_mTOR} and \ref{fig:roc_tgf}, the dashed line %
demonstrates the random %
classifier, which can be evaluated as the worst case. %
In Fig. \ref{fig:roc_mTOR}, five ROC curves for $10$, $20$, $30$, $40$, and $50$ %
mTOR training samples were presented. On the other hand, for the TGF-$\beta$ dataset %
in Fig. \ref{fig:roc_tgf}, the ROC curves were plotted for only $10$, $20$, %
and $30$ training %
samples since the improvements in the results for higher training sample counts %
are not significant. One can easily observe from Fig. \ref{fig:roc_mTOR} %
and \ref{fig:roc_tgf} that the AUC values %
increase consistently while the number of training samples grows for both %
datasets. %

\begin{table}[h!]
\caption{Classifier performance metrics for imbalanced mTOR and TGF-$\beta$ datasets. The number of training samples are presented on headline.}
\centering
\begin{tabular*}{.47\textwidth}{@{}llcccc@{}}
\toprule
& \textbf{Patients} / $\textbf{Controls}$ & $\mathbf{10/40}$ & $\mathbf{20/80}$ & $\mathbf{30/120}$ & $\mathbf{40/160}$ \\
\midrule
\multirow{4}{*}{\rotatebox[origin=c]{90}{mTOR}}%
& True Neg. Rate & $0.8121$ & $0.9003$ & $0.9333$ & $0.9540$ \\
& True Pos. Rate & $0.9990$ & $0.9991$ & $0.9993$ & $0.9991$ \\
& Recall         & $0.0736$ & $0.5560$ & $0.7131$ & $0.8060$ \\
& Specificity    & $0.9999$ & $0.9999$ & $0.9999$ & $0.9998$ \\
& MCC            & $0.2284$ & $0.7062$ & $0.8147$ & $0.8759$ \\
\midrule
\multirow{4}{*}{\rotatebox[origin=c]{90}{TGF-$\beta$}}%
& True Neg. Rate & $0.8236$ & $0.9302$ & $0.9672$ & $0.9813$ \\
& True Pos. Rate & $0.9992$ & $0.9994$ & $0.9993$ & $0.9996$ \\
& Recall         & $0.1396$ & $0.6990$ & $0.8640$ & $0.9233$ \\
& Specificity    & $0.9999$ & $0.9999$ & $0.9999$ & $0.9999$ \\
& MCC            & $0.4414$ & $0.8054$ & $0.9136$ & $0.9515$ \\
\bottomrule
\end{tabular*}
\label{tab:tab2}
\end{table}
\begin{figure}[h!]
\begin{center}
\includegraphics[width=.5\textwidth]{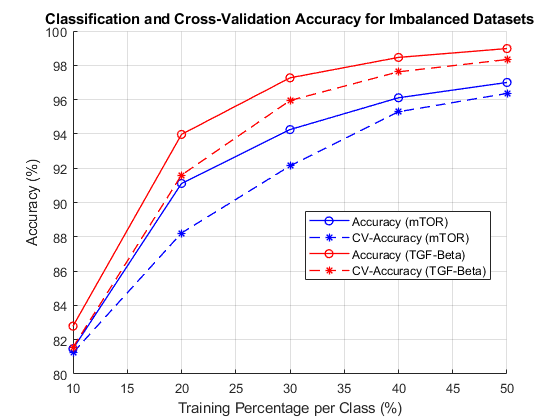}
\end{center}
\caption{Average overall and cross-validation accuracies for varying %
training percentages using imbalanced datasets.}
\label{fig:imbalancedAccuracies}
\end{figure}

In addition to previous analyses, it is also crucial to investigate the %
performance of the proposed method for imbalanced datasets. To this end, two new datasets %
were created from the existing ones for mTOR and TGF-$\beta$ networks. These datasets contain %
$100$ patient and $400$ control samples. For each dataset, we selected random networks %
from both groups as training samples by following the fixed training ratios of $10\%$, $20\%$, $30\%$, $40\%$ and $50\%$, respectively. Thus, the training sample amounts for patient and %
control groups are determined as $10/40$, $20/80$, $30/120$, $40/160$, and $50/200$, %
respectively. That means, in the imbalanced datasets, the number of training control %
networks are fixed is four times the number of training patient samples.

In Fig. \ref{fig:imbalancedAccuracies}, it is shown that the calculated OA values %
at $10\%$ training ratio for imbalanced mTOR and TGF-$\beta$ datasets are approximate%
ly $82\%$ and $83\%$, respectively. These values are less than the presented %
OAs in Fig. \ref{fig:accuracy} for the same training percentage. Moreover, in Fig. \ref{fig:imbalancedAccuracies}, the %
gaps between the OAs and CV accuracies for both datasets at %
a $10\%$ training percentage %
are close, in contrast with the findings in Fig. \ref{fig:accuracy}. On the other %
hand, these gaps tend to shrink after $20\%$ training ratio consistently which are %
similar to the observations given in Fig. \ref{fig:accuracy}. The OA values and CV %
accuracies for both dataset increase constantly. The OA values at a $50\%$ training %
ratio %
for imbalanced mTOR and TGF-$\beta$ datasets are calculated as $97\%$ and $99\%$, %
respectively. These values are in harmony with the accuracies %
calculated for the balanced datasets and provided in Fig. \ref{fig:accuracy}.

To analyse the stability and performance of the proposed method for %
imbalanced datasets, the relevant machine learning metrics for both %
mTOR and TGF-$\beta$ are calculated. These values are presented in Table %
\ref{tab:tab2}, but the results for $50\%$ training rate are not provided due to the limited space. According to Table \ref{tab:tab2}, the recall values %
at $10$ training rate for both imbalanced datasets are quite low. That means the proposed %
classification scheme struggles to predict the patient samples correct%
ly by employing $10$ random patient features. On the other hand, the %
recall values tend to increase rapidly as the number of training patient %
samples grow. The recall values for imbalanced datasets underperform %
the recall values for the balanced datasets. The same issue can be %
remarked on for the MCC metric. However, it can be observed that all %
presented metrics approach their maximum, which is $1$, as the number of training samples increases.

\section{Discussion}

In this new age of\lq\lq holistic genetics\rq\rq, most efforts so far have been devoted to identifying specific gene networks \cite{almeida,choobdar,cui,hawe,maiorino,menche}. The attempts to study the behaviour of the variants in these network genes in their context are still few and timid because of the technical difficulties of handling the vast amount of variants between individuals \cite{mellerup2017,papadimitriou2019}.

GWAS studies are efficient in detecting the significant genomic variants for particular phenotypes. This knowledge made it possible to identify the relevant variants for certain diseases and which genomic variants are causal for the predisposition to the disease, giving hope to compare individual variations in gene networks to predict the personal predisposition to diseases. The technique in use to assess an individual’s susceptibility to a particular physical or mental illness is the PRS, which relies on determining the set of the SNPs that were known as risky from other studies including GWAS. However, polygenicity is a significant problem for this technique, as many phenotypic traits are affected by too many genes, making it hard to calculate PRS\cite{choobdar,hawe,maiorino,menche,fang,wangReview,uffelmann2021,konuma2021,khera2018,silberstein2021,weng2011,xie2021,zhu2018}.  
Another difficulty of the PRS is the requirement of knowledge about the weighted effect of each variant on the phenotype. Since with the available techniques, it is impossible to study the combinatorial effects of more than a few variants, new approaches are required if it is desired to assess the effect of all the variants at once.

To be able to interpret the impact and importance of the millions of variants obtained in a single Next Generation Sequencing study, focusing on data in terms of patterns and corresponding less dimensional entities is rational.

Here, we propose a high dimensional modelling based method to analyse all the variants in a gene network together. In our approach, we apply CGR\cite{CGR1,CGR2,CGR3} as a pre-processing tool to convert the sequencing data of the genes in the network to 2-D binary image patterns. Then, these patterns were aligned (as a three-dimensional tensor) in succession, creating a cube. Afterwards, these tensors were decomposed and represented in terms of their less dimensional features using EMPR. Finally, SVM, which is a multi-class classification algorithm, was fed with three EMPR vector features for each network.

To effectively assess the discrimination ability of our approach, we chose to test it on synthetic datasets. We generated sample patient and control datasets prepared from the reference sequences of the mTOR and TGF-$\beta$ networks of $31$ and $93$ genes of different sizes, respectively. Our findings revealed an accuracy higher than $96\%$ employing only $50$ training features out of $400$ data samples from both control and patient groups. The AUC results indicate that the proposed classifier’s performance in distinguishing between two classes is admirable. Consequently, our results indicate that the proposed CGR, EMPR, and SVM ensemble provides efficient classification performance. 

One of the strengths of our approach is its capability to handle data of various sizes. It is independent of the length of the sequences and the number of genes in the networks. It can be easily applied to all gene networks and is an easy-to-implement algorithm. Also, -unlike PRS- it does not need predetermined knowledge of which variants are relevant and how much they have an impact. The only necessity is to know the relevant gene network. After that, it utilizes the raw sequence data from the case and normal subjects and determines the patient and normal CGR patterns according to the particular variant combinations. 

Since human has a diploid genome and each variant in the human genome has a zygosity (homozygous, heterozygous, or hemizygous) state, this method could be considered challenging for human variant data. In addition, there are two positional possibilities (cis and trans) regarding
any two variants at a heterozygous state. However, for a gene network, the positional state of the variants in the network genes is not important because each gene works as a separate unit of the network. Therefore, the positional state of the variants between different genes is not a limitation of our method. On the other hand, the positional state of the variants within the same gene may become a limitation because each one of two heterozygous variants in a gene may be located in the same or different protein molecule. To overcome this limitation, our method may require some modifications to be applied in the diploid case. These modifications may include the representation of each base substitution with IUPAC codes (R for A/G, S for C/G, W for A/T, M for A/C, for instance) as additional features or properties to CGR rules. Considering these additional features, the CGR process may be updated. Thus, a 4-D sample space may occur and the orthogonality of the sample
space is preserved. Finally, EMPR, which is suitable for $N$–D structures may be implemented to extract the features of the network under consideration.

To the best of our knowledge, our proposed approach to decipher the outcomes of gene networks based on specific combinations of all the variants in the module is original and unique. The observations and findings in this study encourage us that our approach has the potential to be
a diagnostic tool as well as determines individual disposition to polygenic multifactorial conditions.

In addition, comparative studies may be conducted from an evolutionary perspective. This study may also be adapted to different scientific fields, e.g. population genetics, phylogenetics, advanced genomics studies, etc. Furthermore, provided that the necessary fieldwork is done, this method can also be used in talent determination, thus providing the opportunity to receive appropriate training
from an early age.

\section{Conclusion}

According to our results and observations, using high-%
dimensional computational modelling for gene network %
and network-specific gene variant analyses in a holistic manner seems rational and reliable. Our promising %
results encourage us to perform the proposed approach on diploid sequence data for more comprehensive %
future studies. %


\section*{Acknowledgments}

The authors would like to thank Osman \"Ozkan %
for language editing.

\ifCLASSOPTIONcaptionsoff
  \newpage
\fi



\bibliographystyle{IEEEtran}
\bibliography{IEEEabrv,references}
%

%
\vskip-1cm
\begin{IEEEbiography}
[{\includegraphics[width=1in,height=1.25in,clip,keepaspectratio]{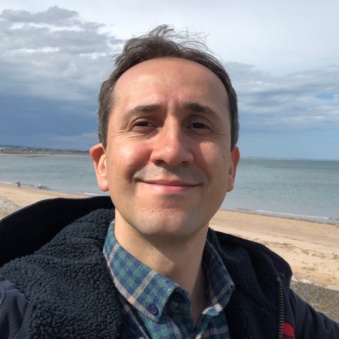}}]{Suha Tuna}
received a Ph.D. degree in computational science and engineering from Istanbul Technical University (ITU), Istanbul, Turkey, in 2017. He is an assistant professor with the Department of Computational Science and Engineering at the Informatics Institute, ITU. His research interests cover  high dimensional modeling, high performance computing, hyperspectral imagery, bioinformatics and machine learning.
\end{IEEEbiography}
\vskip-1cm
\begin{IEEEbiography}
[{\includegraphics[width=1in,height=1.25in,clip,keepaspectratio]{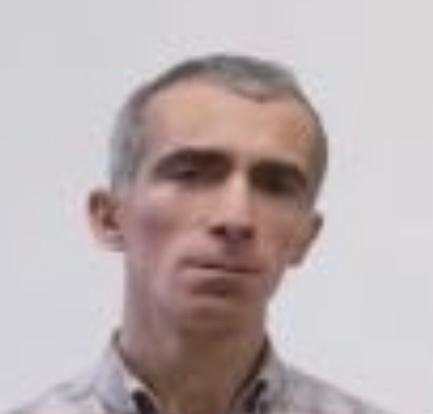}}]{Cagri Gulec}
received his BSc in Biomedical Sciences from Istanbul University, Cerrahpaşa Medical Faculty, and MS. and Ph.D. degrees in Genetics from Istanbul University, Institute of Health Sciences, Istanbul, Turkey. He is currently working at Istanbul University, Istanbul Medical Faculty, Department of Medical Genetics. His research interests include the molecular basis of genetic diseases and bioinformatics..
\end{IEEEbiography}
\vskip-1cm
\begin{IEEEbiography}
[{\includegraphics[width=1in,height=1.25in,clip,keepaspectratio]{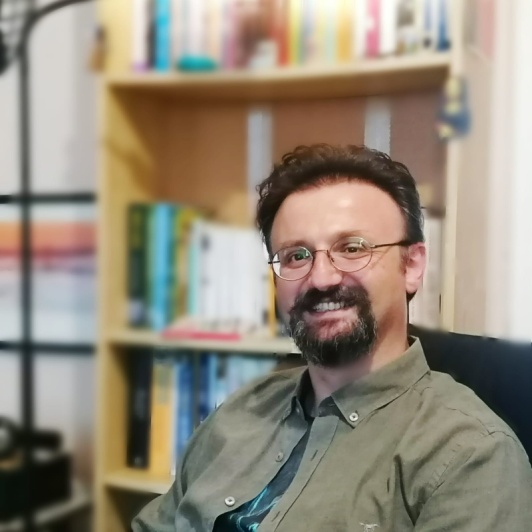}}]{Emrah Yucesan}
received his BSc. in Biomedical Sciences from Istanbul University, Cerrahpaşa Medical Faculty, and his M.S. and Ph.D.degrees in Genetics from Istanbul University, Institute of Health Sciences, Istanbul, Turkey. Emrah Yucesan got his associate professor title in Medical Genetics at 2021. He is currently working at Istanbul University-Cerrahpasa, Institute of Neurological Sciences, Department of Neuroscience. His research interests include neurogenetics and rare diseases. He also interests in bioinformatics and conducts several studies.
\end{IEEEbiography}
\begin{IEEEbiography}
[{\includegraphics[width=1in,height=1.25in,clip,keepaspectratio]{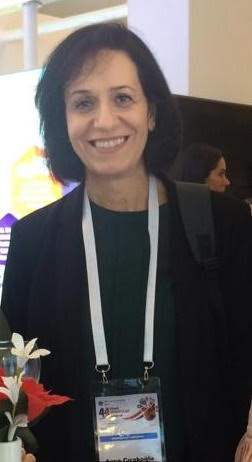}}]{Ayse Cirakoglu}
received her BSc. in Biomedical Sciences from Istanbul University, Cerrahpaşa Medical Faculty, and M.S. and Ph.D. degrees in Genetics from Istanbul University, Institute of Health Sciences, Istanbul, Turkey. She is currently working as Associate Professor at the Department of Medical Biology, Cerrahpaşa Medical Faculty. Her research interests include cytogenetics, molecular cytogenetics, cancer genetics, epigenetics, and gene network analysis.
\end{IEEEbiography}
\newpage
\begin{IEEEbiography}
[{\includegraphics[width=1in,height=1.25in,clip,keepaspectratio]{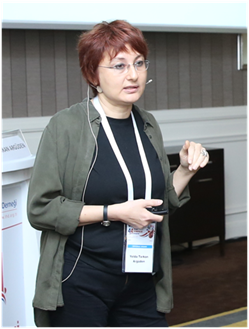}}]{Yelda Tarkan Arguden}
graduated from the Department of Biomedical Sciences, Cerrahpaşa Faculty of Medicine, Istanbul University, in 1988. She received her MSc. in Medical Genetics in 1991 and a Ph.D. in Genetics in 1999 from the Institute of Health Sciences, Istanbul University. She is currently working as Associate Professor at the Medical Biology Department of Cerrahpaşa Faculty of Medicine, Istanbul University-Cerrahpaşa. Her research interests include cytogenetics, cancer cytogenetics, epigenetics, and gene network analysis.
\end{IEEEbiography}





\end{document}